\documentclass[conf]{new-aiaa}
\usepackage[utf8]{inputenc}

\usepackage{graphicx}
\usepackage{amsmath}
\usepackage[version=4]{mhchem}
\usepackage{siunitx}
\usepackage{subcaption}
\usepackage{longtable,tabularx}
\usepackage{booktabs}
\usepackage{ifthen}
\usepackage{xcolor}

\newboolean{colortext}
\setboolean{colortext}{false} % Set to 'true' for blue text, 'false' for black

\newcommand{\bt}[1]{%
    \ifthenelse{\boolean{colortext}}%
        {\textcolor{blue}{#1}}% if colortext is true
        {#1}% if colortext is false
}

\title{Learning and Autonomy for Extraterrestrial Terrain Sampling: An Experience Report from OWLAT Deployment}

\author{Pranay Thangeda\footnote{Graduate Student, Department of Aerospace Engineering, pranayt2@illinois.edu}, Yifan Zhu\footnote{Graduate Student, Department of Computer Science, yifan16@illinois.edu}, Kris Hauser\footnote{Professor, Department of Computer Science,  kkhauser@illinois.edu}, and Melkior Ornik\footnote{Assistant Professor, Department of Aerospace Engineering, mornik@illinois.edu}}
\affil{University of Illinois Urbana-Champaign, Urbana, Illinois 61801}
\author{Ashish Goel\footnote{Robotics Technologist, Robotic Surface Mobility Group, ashish.goel@jpl.nasa.gov}, Erica L Tevere\footnote{Robotics Technologist, Robotic Surface Mobility Group, erica.l.tevere@jpl.nasa.gov}, Adriana Daca\footnote{Robotics Technologist, Robotic Surface Mobility Group, adriana.daca@jpl.nasa.gov}, and Hari D Nayar\footnote{Member of Technical Staff, Robotic Surface Mobility Group, hdnayar@jpl.nasa.gov}}
\affil{Jet Propulsion Laboratory, California Institute of Technology, Pasadena, California 91109}
\author{Erik Kramer\footnote{Graduate Student, Department of Mechanical and Aerospace Engineering, ehkramer@ucla.edu}}
\affil{University of California, Los Angeles, Los Angeles, California 90095}

\begin{document}

\maketitle

\begin{abstract}
Extraterrestrial autonomous lander missions increasingly demand adaptive capabilities to handle the unpredictable and diverse nature of the terrain. This paper discusses the deployment of a Deep Meta-Learning with Controlled Deployment Gaps (CoDeGa) trained model for terrain scooping tasks in Ocean Worlds Lander Autonomy Testbed (OWLAT) at NASA Jet Propulsion Laboratory. The CoDeGa-powered scooping strategy is designed to adapt to novel terrains, selecting scooping actions based on the available RGB-D image data and limited experience. \bt{The paper presents our experiences with transferring the scooping framework with CoDeGa-trained model from a low-fidelity testbed to the high-fidelity OWLAT testbed. Additionally, it} validates the \bt{method's} performance in novel, realistic environments, and shares the lessons learned from deploying learning-based autonomy algorithms for space exploration. Experimental results from OWLAT substantiate the efficacy of CoDeGa in rapidly adapting to unfamiliar terrains \bt{and} effectively making autonomous decisions under considerable domain shifts, thereby endorsing its potential utility in future extraterrestrial missions.
\end{abstract}

\section{Introduction}

The exploration of ocean worlds stands as a pivotal element in \bt{humanity's} exploration of our solar system, encompassing critical research objectives including the quest for potential signs of life and the comprehensive understanding of conditions fostering habitability  \cite{oceanworldroadmap}, \cite{howell2021ocean}, \cite{europaprogram}. \bt{Robotic exploration missions are essential for the exploration of potentially habitable ocean worlds}. Past lander and rover missions including the Mars exploration program \cite{mep} and the Perseverance rover mission \cite{perseverance} are human-in-the-loop systems with expert teams on Earth supervising the terrain sampling process and controlling them based on the collected data. However, unlike Mars missions, many of the ocean world missions, including the Europa Lander mission concept \cite{europareport}, are anticipated to have short durations, on the order of tens of days, due to the intensity of the radiation environment, adverse thermal conditions, low availability of solar energy, and using battery as the sole power source. The limited mission duration combined with the long communication delays between Earth and the ocean worlds necessitates a high degree of autonomy for the \bt{lander's success} \cite{autonomyow}. 

\bt{The Europa lander's primary objectives include collecting terrain samples for in situ analysis of surface and sub-surface materials.} Autonomy in terrain sampling missions is challenging due to the high degree of uncertainty in the surface topology at the landing site, terrain material properties, composition, and appearance. Constraints on the number of samples that can be analyzed in-situ, coupled with the risk of system failures, further limits the extent of exploration \cite{thangeda2022adaptive}. Any realistic sampling strategy needs to be able to make decisions under uncertainty and rapidly adapt to the environment in order to maximize the scientific return from the mission. To address these challenges, recent efforts have focused on the development of capabilities for autonomous excavation site selection \cite{ono2020ariel}, \cite{zhu2023codega}, autonomy software prototype to execute complex and highly constrained missions with limited human intervention \cite{wagner2023demonstrating}, and field testing the functional autonomy stack in terrestrial analog environments \cite{bowkett2023demonstration}. Specifically, our prior work~\cite{zhu2023codega} proposed Deep Meta-Learning with Controlled Deployment Gaps (CoDeGa), an adaptive scooping strategy that uses deep Gaussian processes trained with a novel meta-learning approach. CoDeGa learns online from very limited experience on target terrains despite large domain shifts from the training set. These prior experiments were conducted on a low-fidelity testbed at the University of Illinois Urbana-Champaign (UIUC) designed to rapidly collect large-scale data for training and testing the models on a wide range of terrains.

Building upon this foundational work, the next crucial step is to validate and refine our strategy within a more sophisticated and representative environment. The Ocean Worlds Lander Autonomy Testbed (OWLAT) \cite{nayar2021development} at NASA Jet Propulsion Laboratory (JPL) provides such an environment. OWLAT is a high-fidelity testbed developed to validate autonomy algorithms for future ocean world missions. It serves as a state-of-the-art platform for simulating various potential future planetary missions over a wide range of dynamic environments, including surface operations on small bodies where recreating the dynamics in low gravity is critical. By integrating CoDeGa with OWLAT, we aim to assess the robustness of the proposed autonomy algorithms in conditions that closely mimic those of actual extraterrestrial landscapes, thereby bridging the gap between preliminary tests and real-world deployment.

In this paper, we report our experiences deploying the CoDeGa-trained adaptive scooping model on the OWLAT testbed. Our contribution is threefold: (1) we assess the feasibility of transferring the model across systems with similar sensor suites, end-effectors, and primitive actions, (2) we validate the model's adaptability in novel, realistic environments and (3) we share the lessons learned from designing and deploying learning-based autonomy algorithms in the context of space exploration. Experimental results provide strong evidence that CoDeGa-trained model adapts to the significant domain shifts presented by the OWLAT testbed, reinforcing its applicability and promising role in autonomous terrain sampling for future off-world missions.

\section{Preliminaries and Background}
This section describes the scooping problem setting in detail and provides a brief overview of the solution approach using the CoDeGa-trained model as proposed in \cite{zhu2023codega}. We also highlight the differences between the UIUC testbed and OWLAT and the importance of deploying and validating the solution approach on the OWLAT testbed.

\subsection{Scooping Problem}
We study the problem of scooping in which the goal is to collect high-volume samples from the lander's workspace with a limited budget of attempts. The problem is formulated as a sequential decision-making problem where the robot observes the terrain RGB-D image $o \in \mathcal{O}$, uses a \textit{scooping policy} to apply action $a \in \mathcal{A}(o)$ where $\mathcal{A}(o)$ is a discrete set of parameterized scooping motions dependent on the RGB-D image, and receives reward $r \in \mathcal{R}$, which is the scooped volume.

Given a target terrain $T_*$, the robot's goal is to find a series of scooping motions that maximize the total reward from scoops across the first $k$ attempts. During the $n$-th attempt, for $n \leq k$, the robot has access to the history of scoops on this terrain $H = \{(o^j,a^j,r^j)\,|\,j=1,\ldots,n-1\}$.  The robot also has access to prior scooping experience, which consists of a set of $M$ terrains $\{T_1, \dots, T_M \}$, and a training dataset $D_i = \{(o^j,a^j,r^j)\,|\,j=1,\ldots,N_i\}$ of past scoops and their rewards for each terrain $i=1,...,M$.

We suppose that a \textit{latent variable} $\alpha$ characterizes a given terrain's indirectly observed properties including composition, material properties, and topography. Let $\alpha_*$ characterize terrain $T_*$ and $\alpha_i$ characterize terrain $T_i$ for $i=1,\ldots,M$. The observation depends on $\alpha$ and action rewards are unknown functions of both action and $\alpha$. Standard supervised learning for modeling $r\approx f(o,a)$ is effective when $\alpha$ aligns with training terrains and is deducible from observation $o$, or if rewards are not heavily tied to unobserved latent characteristics. Performance drops when $T_*$ is outside the training distribution or when observation $o$ inadequately indicates the terrain's latent aspects affecting rewards.

\bt{We propose an adaptive online learning strategy for better sampling performance on a terrain $T_*$ despite significant differences from training terrains. This method utilizes real-time data from $H$, adapting to terrain $T_*$ and countering domain shifts from the training set.}

\subsection{Adaptive Scooping using CoDeGa-trained Model}

\begin{figure}[hbt!]
    \centering
    \includegraphics[width=1.0\textwidth]{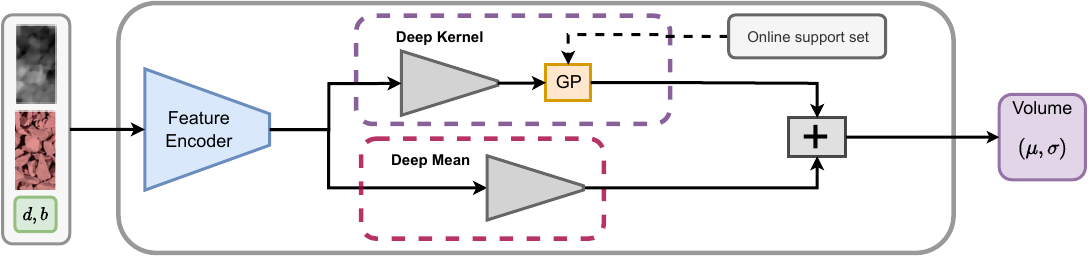}
    \caption{Overview of the CoDeGa-trained model with deep mean, deep kernel, and feature encoder modules.}
    \label{fig:codega_model}
\end{figure}

The solution approach proposed in \cite{zhu2023codega}, Deep Meta-Learning with Controlled Deployment Gaps (CoDeGa), leverages a deep Gaussian process (GP) model to capture the relationship between the observation-action pair $(o,a)$ and the reward $r$. The deep Gaussian process model employs deep mean functions and deep kernels where the input to the GP kernel is transformed by a neural network. We note that, in addition to the observation-action pair, the model is also conditioned on the \textit{online support set} that contains the history of previous scoops and their outcomes on the same terrain. In the context of a scooping task, this model predicts the scooped volume given an input consisting of a local patch of an RGB-D image and action parameters, as illustrated in Fig. \ref{fig:codega_model}.

In the CoDeGa training procedure, the training terrains are split into mean and kernel training sets containing different materials. Doing so encourages the kernel to encounter residuals representative of those in out-of-distribution tasks. The deep mean is first trained on the mean training set to minimize error, and the GP kernel is then trained on the residuals of the deep mean model applied to the kernel training set. This process is repeated, similar to $k$-fold cross-validation, with a common kernel trained over aggregated losses across folds. The strength of CoDeGa lies in its ability to generate a model that performs well under deployment gaps, which are common in real-world applications. 

Given a model to robustly predict the scooped volume, \cite{zhu2023codega} employs a Bayesian optimization approach for selecting the scooping action. Rather than simply using the mean prediction of the model, it utilizes an acquisition function that also takes uncertainty into account. This function serves as a scoring system that guides the selection of actions and encourages the exploration of actions with uncertain outcomes, allowing for a more robust performance under varying conditions.

The CoDeGa-based scooping strategy, while effective, requires a large amount of data across different terrains for training the deep mean and kernel. In the next section, we describe the UIUC testbed designed to collect such data with minimal human supervision and the different terrains and compositions on which the data is collected. 

\subsection{UIUC Testbed}
\begin{figure}[hbt!]
    \centering
    \includegraphics[width=.9\textwidth]{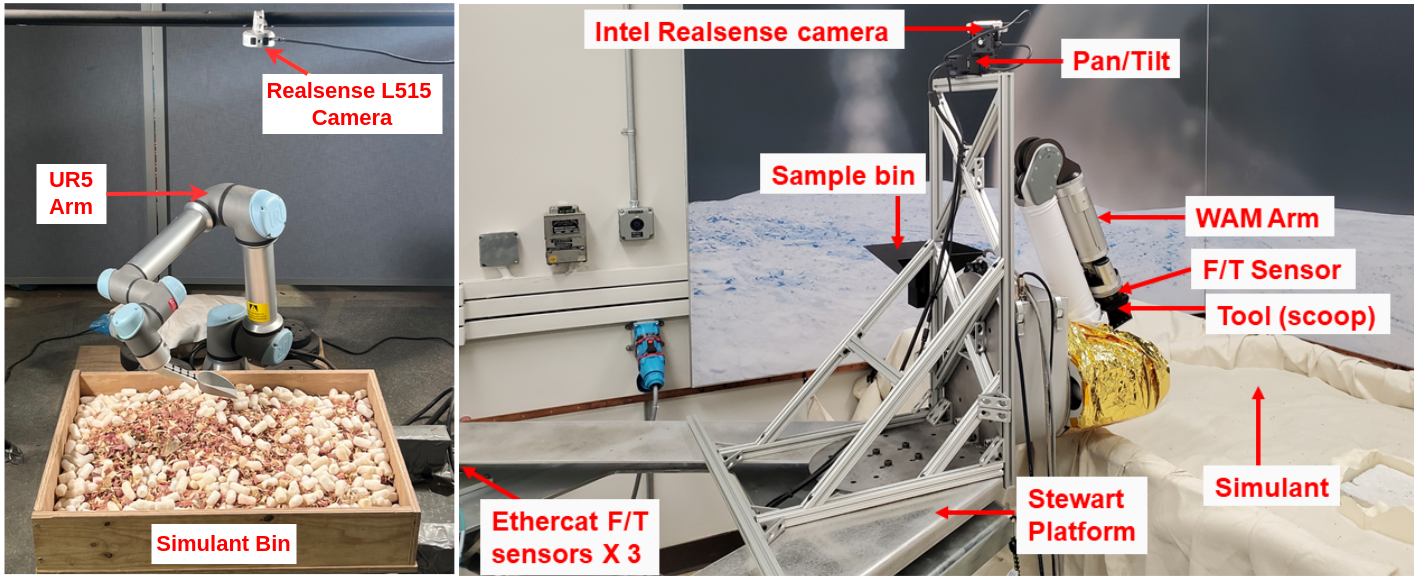}
    \caption{Comparison of the UIUC testbed (left) and the Ocean Worlds Lander Autonomy Testbed (OWLAT) (right).}
    \label{fig:setup_comparison}
\end{figure}

The data collection testbed at the University of Illinois Urbana-Champaign (UIUC) is designed for large-scale data collection and testing of learning-based approaches for scooping tasks. The setup includes a UR5e arm with a scoop mounted on the end-effector, an overhead Intel RealSense L515 RGB-D camera, and a wheeled simulant bin that is approximately 0.9 m x 0.7 m x 0.2 m. A \textit{scoop action} in this setup is a parameterized trajectory for the scoop end-effector tracked by an impedance controller. Fig. \ref{fig:setup_comparison} shows an illustration of the testbed.

A \textit{terrain} is defined as a unique composition of one or more materials. The setup uses multiple wheeled simulant bins to parallelly collect data in one bin and set up a new terrain in another that can be quickly swapped. For each new terrain, data is collected by executing different scooping actions and measuring the volume of scooped material. A detailed description of the different materials considered and the scoop action parameters is provided in \cite{zhu2023codega}. 

The UIUC testbed enables rapid data collection at the cost of limited representation of the \bt{ocean world lander system complexity}. In the next section, we describe the OWLAT testbed designed for validating autonomy algorithms for ocean worlds missions.

\subsection{Ocean Worlds Lander Autonomy Testbed}
The Ocean Worlds Lander Autonomy Testbed (OWLAT) \cite{nayar2021development} is a testbed developed at NASA Jet Propulsion Laboratory (JPL) to test and validate the performance of various autonomy algorithms and architectures for future missions to ocean worlds. As shown in Fig. \ref{fig:setup_comparison}, the testbed hardware consists of a 7-DOF Barrett WAM7 robotic arm with a host of interchangeable end-effector tools representing the manipulator, an intel Realsense D415 mounted on a pan-tilt mount for 3D perception, and force-torque sensors located at the interface between the arm and the platform and also at the end of the arm's wrist. The testbed also has a 6-DOF Stewart platform that is used to simulate the lander and a simulant area that hosts the testbed's terrains with different simulants and surface features. 

OWLAT hardware is complemented by a software interface to command the manipulator and the camera along with its pan-tilt mount for carrying out surface operations and collecting data. The force-torque sensors located at the end of the wrist and at the interface between the arm and the Stewart platform play a critical role in replicating the dynamical environment such landers are likely to experience on the low-gravity icy moons of Jupiter and Saturn. As the tool interacts with the simulant in the testbed, the reaction forces measured are fed into a dynamics model of the system. The computed motion is imposed on the Stewart platform in real-time. The use of high bandwidth Ethercat force-torque sensors allows OWLAT to close \bt{the control-sensing loop at 500 Hz} and study test cases demonstrating how interaction with the surface on objects with gravity as low as Enceladus ($g = 0.13 \, \text{m/s}^2$) can cause the legs of the lander to lift off the ground, thereby achieving Earth gravity compensation without the use of suspension cables and gantry mechanisms. 

In the next section, we describe the process of deploying the CoDeGa-trained model on the OWLAT testbed, highlighting the challenges of interfacing a learned model with new hardware configurations vastly different from the training setting.

\section{Deployment}
This section details the process of deploying the CoDeGa-trained model on OWLAT testbed for selecting scooping actions based on RGB-D information of the robot's workspace and past experience in the deployed environment. 

\begin{figure}[hbt!]
    \centering
    \includegraphics[width=0.95\textwidth]{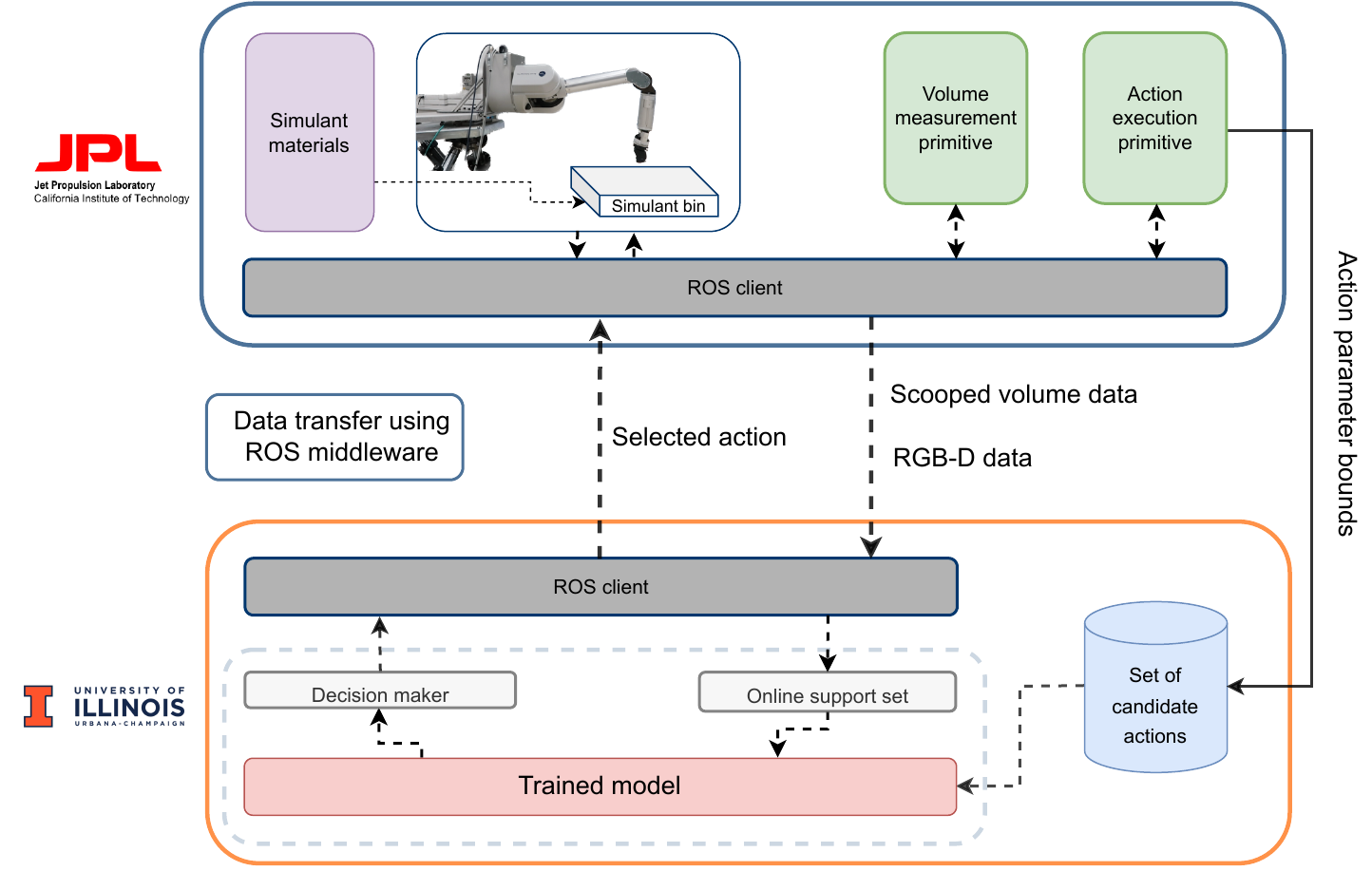}
    \caption{Remote deployment of the adaptive scooping strategy using UIUC-trained model on OWLAT.}
    \label{fig:deployment_overview}
\end{figure}

Deploying the adaptive scooping strategy with the CoDeGa-trained model involves taking a sequence of scooping actions where the \bt{scoop volume estimation model} adapts based on the reward from the previously executed actions. \bt{A single scooping attempt involves the following steps:}

\begin{enumerate}
    \item \bt{collect the RGB-D data capturing the current state of the simulant bin,}
    \item \bt{pass the data and the set of all candidate scoop actions to the client,}
    \item \bt{evaluate the candidate actions using the model, taking into account the history of previous actions and their outcomes,}
    \item \bt{measure the scooped volume, and}
    \item \bt{record the outcome along with the executed action and corresponding observation in the online support set.}
\end{enumerate}

Fig. \ref{fig:deployment_overview} shows an overview of the deployment process with the OWLAT testbed. OWLAT provides a ROS interface with access to the RGB-D camera data and ROS actions over the ROS middleware. We implemented ROS actions that allow us to execute scoop actions parameterized similar to the UIUC testbed. The RGB-D data captured by the camera is passed to the client over the ROS middleware. The deployment client preprocesses the RGB-D data and passes it to the CoDeGa-trained model along with the candidate actions. The decision maker then decides the next action to execute based on the model's evaluations and sends the action to the testbed over the ROS middleware. The testbed executes the action on the robot and measures the scooped volume. Given that the testing on OWLAT involves far fewer overall scooping attempts when compared to the data collection process in the UIUC testbed, we manually measure the scooped volume. We note that a deployment-ready ocean world mission is expected to include the instrumentation to analyze the scientific value of the scooped material, thereby addressing the issue of scooped volume measurement. 

We now discuss in detail the steps involved in processing RGB-D data to be compatible with the CoDeGa-trained model and generating the candidate action set that can be easily implemented on any robotic system with a scoop end-effector.

\subsection{Processing RGB-D Data}

The CoDeGa-trained model utilizes a convolutional neural network-based feature encoder to extract relevant information from the RGB-D data. While effective, these encoders are susceptible to variations in input data, requiring a degree of consistency in feature scale and camera orientation relative to the training set for optimal performance. The UIUC testbed standardizes data capture across materials with varying feature scales by using a static overhead RGB-D camera, thereby streamlining data collection and preprocessing by minimizing environmental inconsistencies.

In contrast, real-world systems rarely offer such controlled conditions. \bt{For example, in the OWLAT testbed, the placement of the RGB-D camera is akin to the camera placement in real-world landers.} It is mounted on a pan-tilt unit at the base of the robot, capturing data from this unique perspective, as shown in Fig. \ref{fig:setup_comparison}. Additionally, real-world data is often further complicated by variations in camera resolution and quality across systems, and it commonly includes noise, occlusions, and other artifacts.

\begin{figure}[hbt!]
    \centering
    \includegraphics[width=1.0\textwidth]{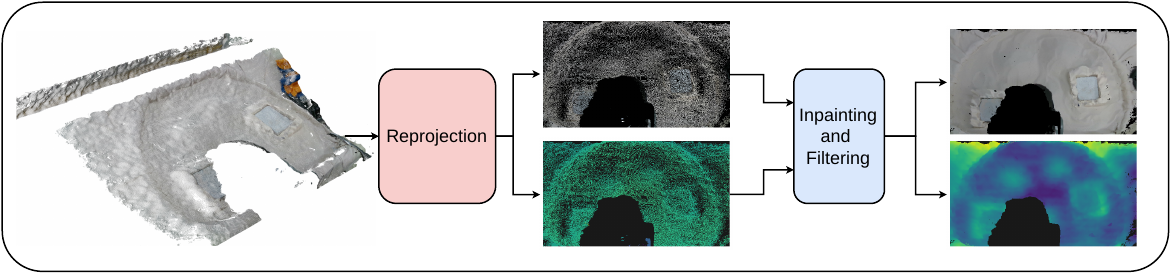}
    \caption{Overview of the steps involved in processing the RGB-D data captured by the OWLAT testbed.}
    \label{fig:vision_pipeline}
\end{figure}

To align the OWLAT data with the training conditions of the CoDeGa-trained model, we developed a preprocessing \bt{pipeline as shown in Fig. \ref{fig:vision_pipeline}}. This pipeline includes (i) reprojecting the point cloud to a top-down view to emulate the UIUC testbed perspective, (ii) \bt{reconstructing missing values to fill} in gaps in the RGB and depth data while accounting for occluded and out-of-range regions, and (iii) using system state and constraints to filter out anomalies. The resulting images are then used to produce localized image patches based on the action parameters. 

In the next section, we discuss how our choice of parameterized motion primitives as actions enables us to easily deploy the CoDeGa-trained model on any robotic system with a scoop end-effector. Additionally, we detail our approach of generating valid candidate actions while accounting for system constraints. 

\subsection{Generating Action Candidates}
\label{sec:action_candidates}
In our implementation, the scoop action is a motion primitive characterized by an initial $x,y$ position, the yaw angle $\theta$ of the scoop, the scooping depth $d$, and \bt{the impedance controller stiffness parameter $b$ that takes a value out of the set \textit{\{low, high\}}.} The z-coordinate is deduced from the depth map at the outset of the action. This action primitive follows the parameterized trajectory described in \cite{zhu2023codega}, which is executed via an impedance controller. By defining the action in the end effector space, we maintain a consistent action interpretation across varied systems. \bt{We adopt the same trajectory for the scoop action primitive as established in \cite{zhu2023codega} for deployment.} Moreover, empirical evidence from \cite{zhu2023codega} indicates that \bt{higher stiffness values for the impedance controller} yield superior performance on diverse terrains; consequently, we fix the stiffness parameter $b$ to \textit{high} value within the OWLAT action primitive to leverage these observed benefits.

\begin{figure}[hbt!]
    \centering
    \includegraphics[width=1.0\textwidth]{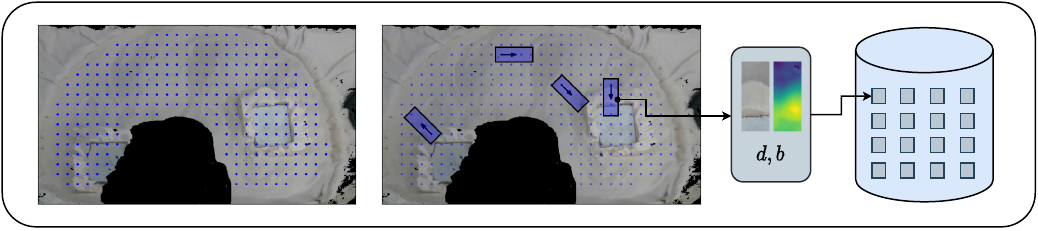}
    \caption{Illustration of the candidate action set generation process using OWLAT testbed data.}
    \label{fig:action_pipeline}
\end{figure}

To synthesize the set of candidate actions, we use the processed depth data and system constraints to delineate regions in the simulant bin amenable to scooping. Within these identified regions, we establish a uniform grid to represent potential action locations. At each grid point, we evaluate eight possible yaw angles—excluding any that are deemed infeasible—coupled with four distinct scooping depths ($d$ values of 0.2 cm, 0.4 cm, 0.6 cm, and 0.8 cm) and a single 'high' stiffness value ($b$). We note that the size of the resultant action set is dynamic, as it adjusts with the changes in the environment. For each candidate action, we generate corresponding RGB and depth image patches centered at the action $x,y$ and aligned with the scoop orientation $\theta$, as illustrated in Fig. \ref{fig:action_pipeline}. The image patches along with the stiffness parameter and the selected depth value are then used to evaluate the action candidate, as illustrated in Fig. \ref{fig:codega_model}.

The next section describes the experiments conducted on the OWLAT testbed to evaluate the performance of the CoDeGa-trained model.

\section{Experimental Evaluation and Discussion}

\begin{figure}[h!]
    \centering
    \begin{subfigure}[b]{0.4\textwidth}
        \centering
        \includegraphics[width=\textwidth]{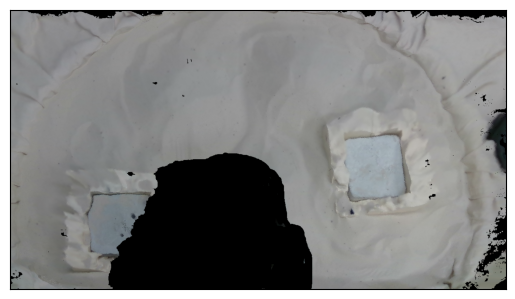}
    \end{subfigure}
    \begin{subfigure}[b]{0.389\textwidth}
        \centering
        \includegraphics[width=\textwidth]{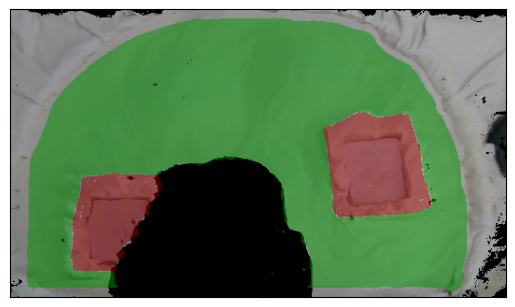}
    \end{subfigure}
    \caption{(left) The terrain composition used for testing in OWLAT testbed. (right) The terrain composition with labels for the \texttt{Regolith} (green) and \texttt{Comet} (red) materials.}
    \label{fig:terrain_labels}
    \end{figure}

We performed the experiments using an active remote connection with the OWLAT testbed operating at the Jet Propulsion Laboratory (JPL), Pasadena, California, and the CoDeGa-trained model deployed for inference at the University of Illinois Urbana-Champaign (UIUC), Champaign, Illinois. Table \ref{tbl:materials} lists all the materials used for training the model along with their corresponding physical properties. For evaluation, we used a terrain designed \bt{by subject matter experts using out-of-distribution materials, \texttt{Comet} and \texttt{Regolith}.} \texttt{Comet} is an unscoopable composition of grey comet simulant material \cite{carey2017development} surrounded by 3D printed PLA features with rugged terrain features from a 3D scan of Devil's Golf Course in Death Valley National Park, painted to match the \texttt{Regolith}'s color. \texttt{Regolith} is a fine sand-like material that is visually distinct from the sand used in training. The two materials were composed together to create a hypothetical representation of the ocean world terrain, as depicted in Fig. \ref{fig:terrain_labels}. 

\begin{table}[h!]
    \begin{center}
    \caption{Materials used in the experiments along with their corresponding grain sizes, categorized by training and testing sets, with U.S. quarter coin for scale.}
    \label{tbl:materials}
    \begin{tabular}{ccccc}
    \toprule
     \multicolumn{4}{c}{Training} & \multicolumn{1}{c}{Testing} \\
     \cmidrule(r){1-4} \cmidrule(l){5-5}
     \shortstack{Sand\\{\scriptsize fine play sand, $<<$ 1 mm}} & 
     \shortstack{Pebbles\\{\scriptsize rocks, 0.8 -- 1.0 cm}} & 
     \shortstack{Slate\\{\scriptsize flat rocks, 2.0--4.0 cm}} & 
     \shortstack{Gravel\\{\scriptsize rocks, 1.5--3.0 cm}} & 
     \shortstack{Comet\\{\scriptsize unscoopable}} \\
     \includegraphics[width=0.145\textwidth]{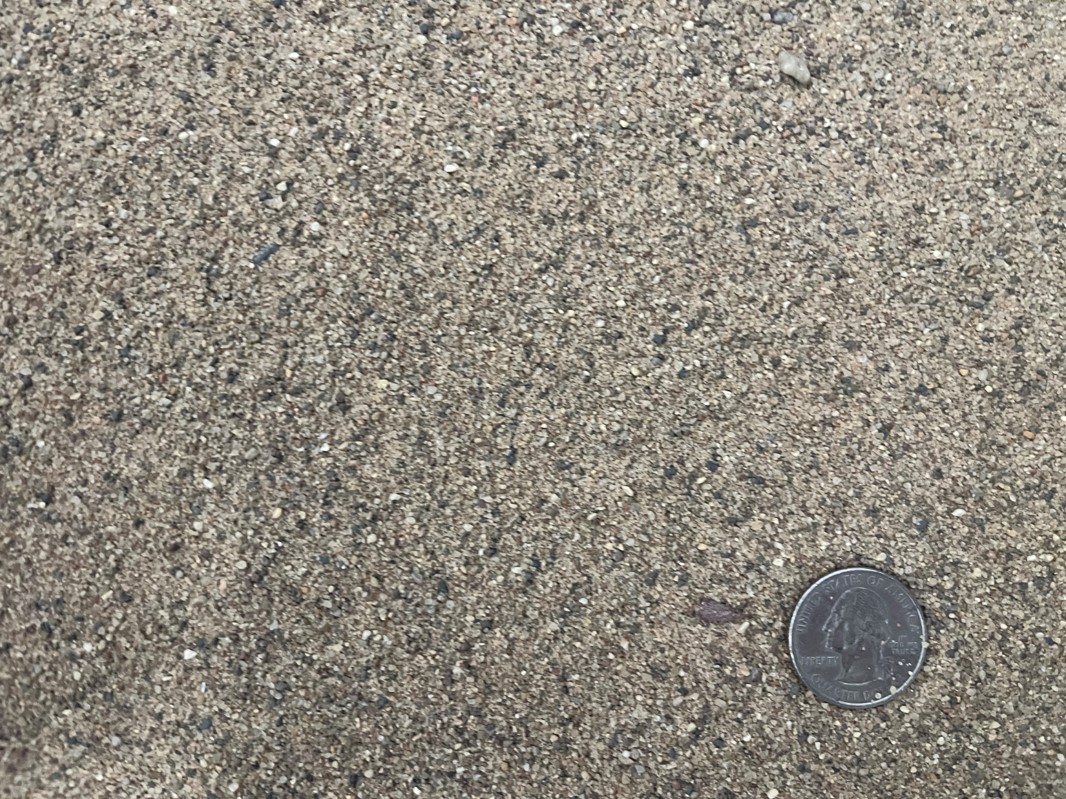} & 
     \includegraphics[width=0.145\textwidth]{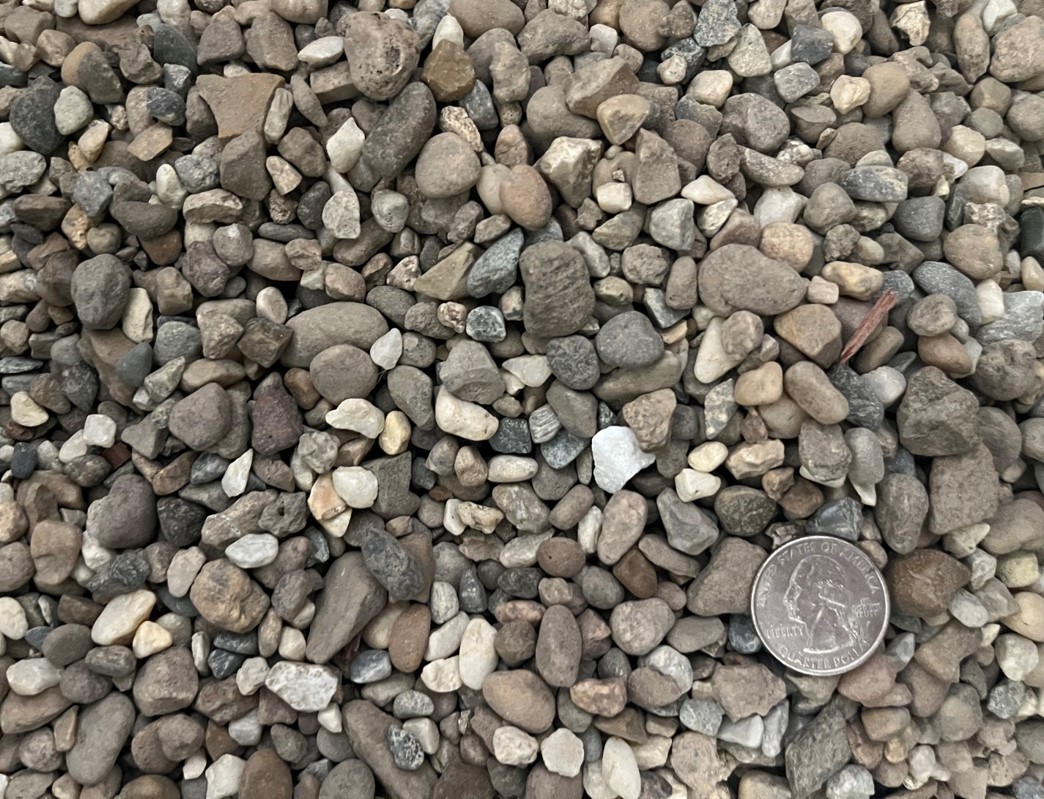} & 
     \includegraphics[width=0.15\textwidth]{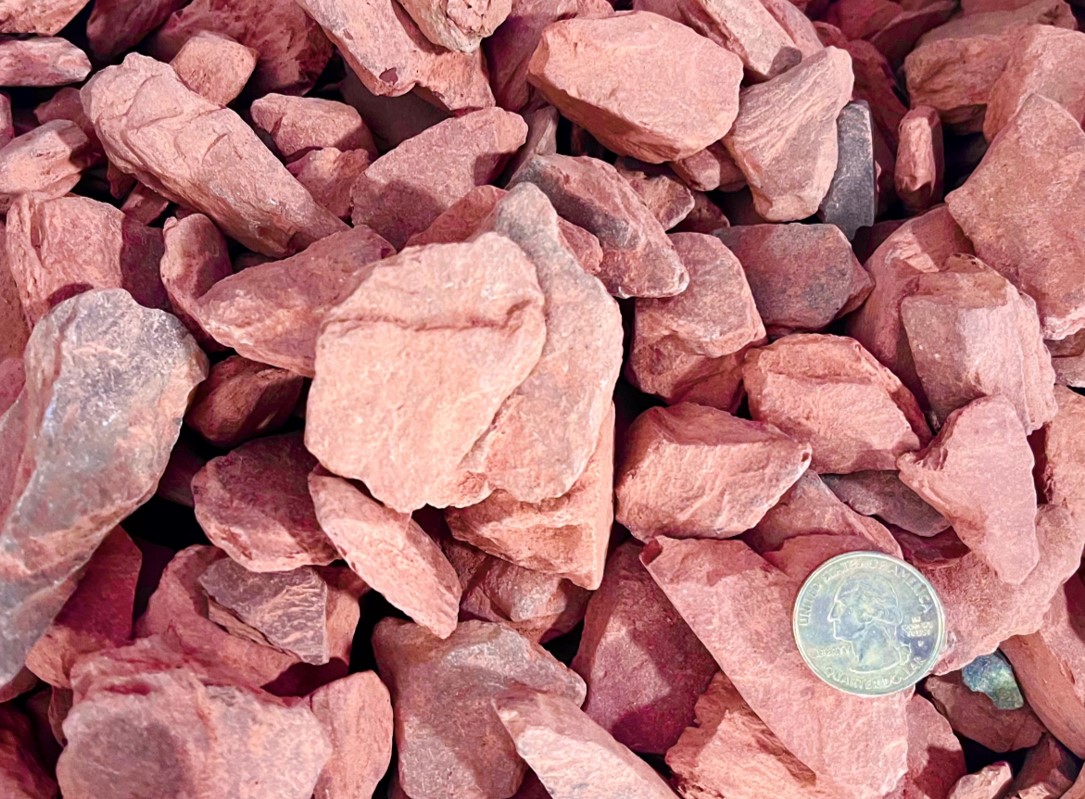} &
     \includegraphics[width=0.15\textwidth]{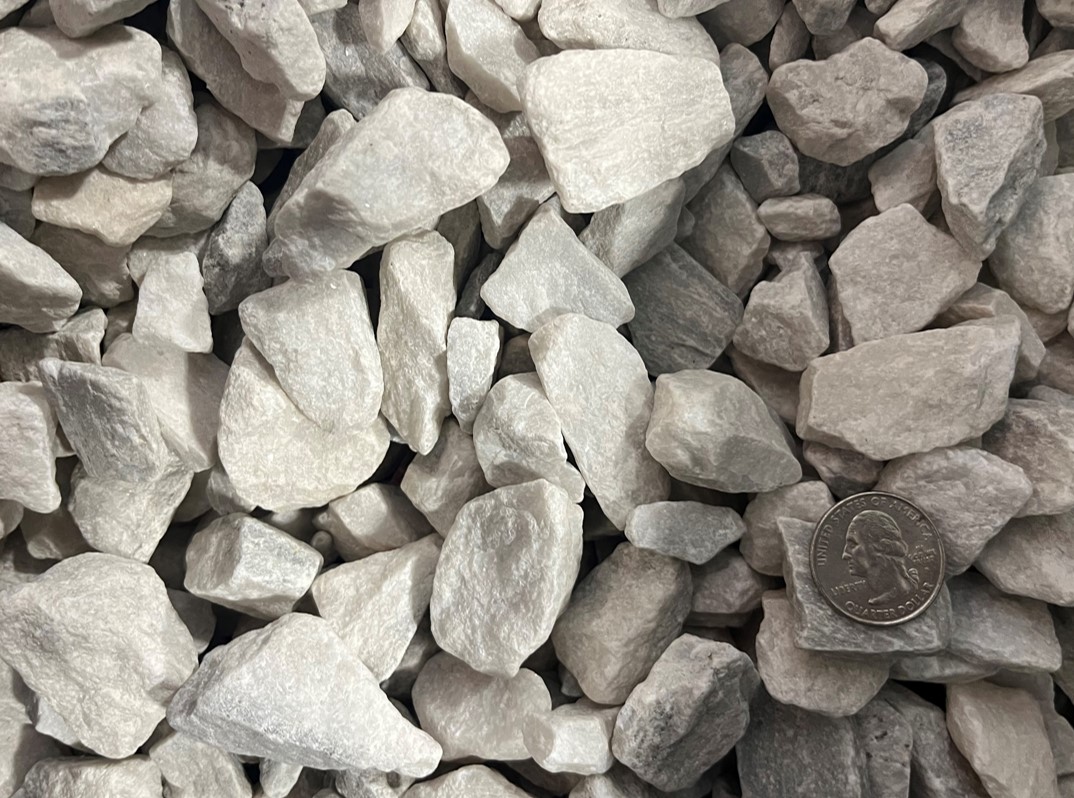} & 
     \includegraphics[width=0.15\textwidth]{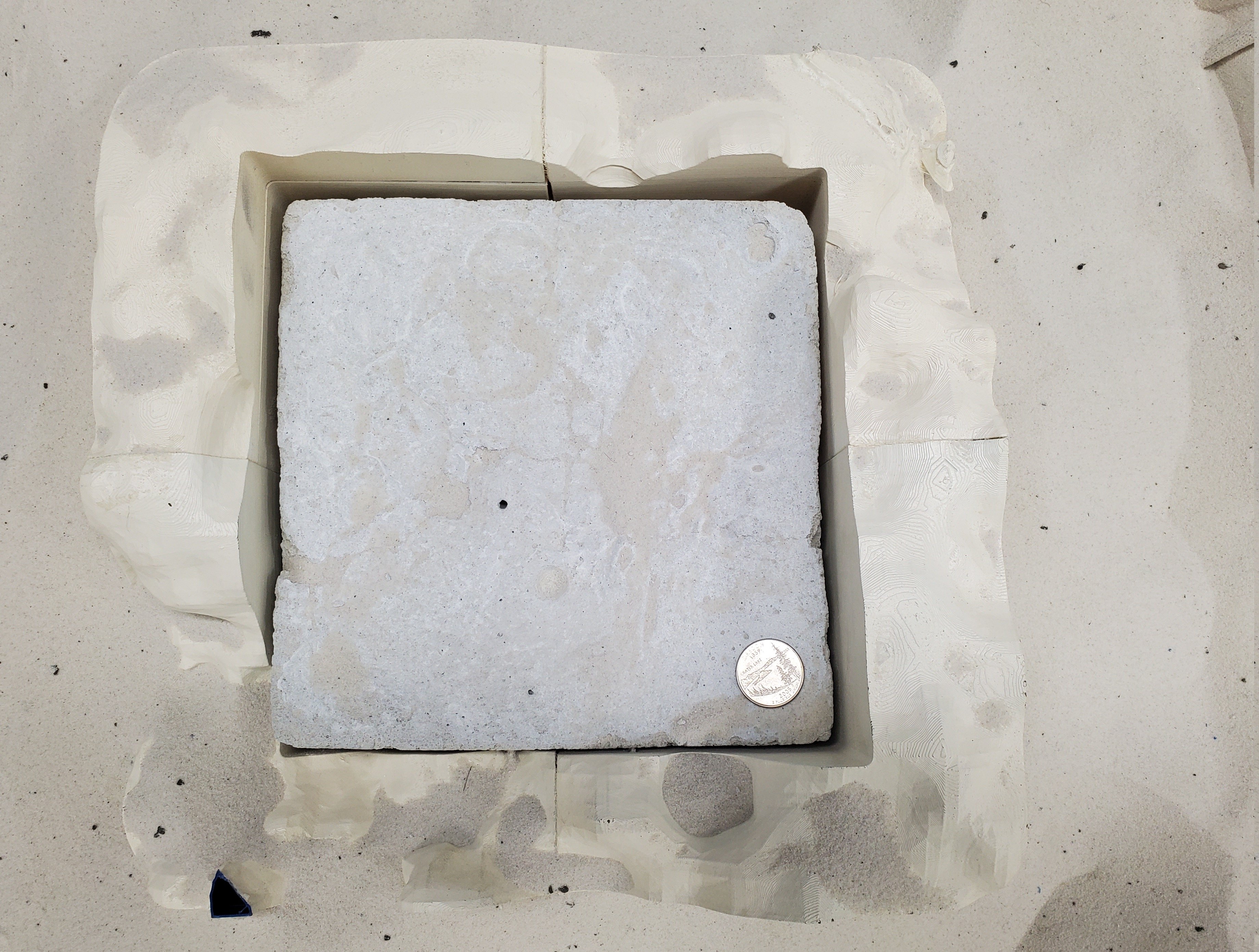} \\
     \cmidrule(r){1-4} \cmidrule(l){5-5}
     \shortstack{Paper Balls\\{\scriptsize crumpled paper, 4.0 -- 6.0 cm}} & 
     \shortstack{Corn\\{\scriptsize dry corn kernels, 0.3--0.7 cm}} & 
     \shortstack{Shredded Cardboard\\{\scriptsize cardboard, 1.0 -- 8.0 cm}} & 
     \shortstack{Mulch\\{\scriptsize red wood landscape mulch}} & 
     \shortstack{Regolith\\{\scriptsize fine sand, 0.1 mm -- 0.5 mm}} \\
     \includegraphics[width=0.15\textwidth]{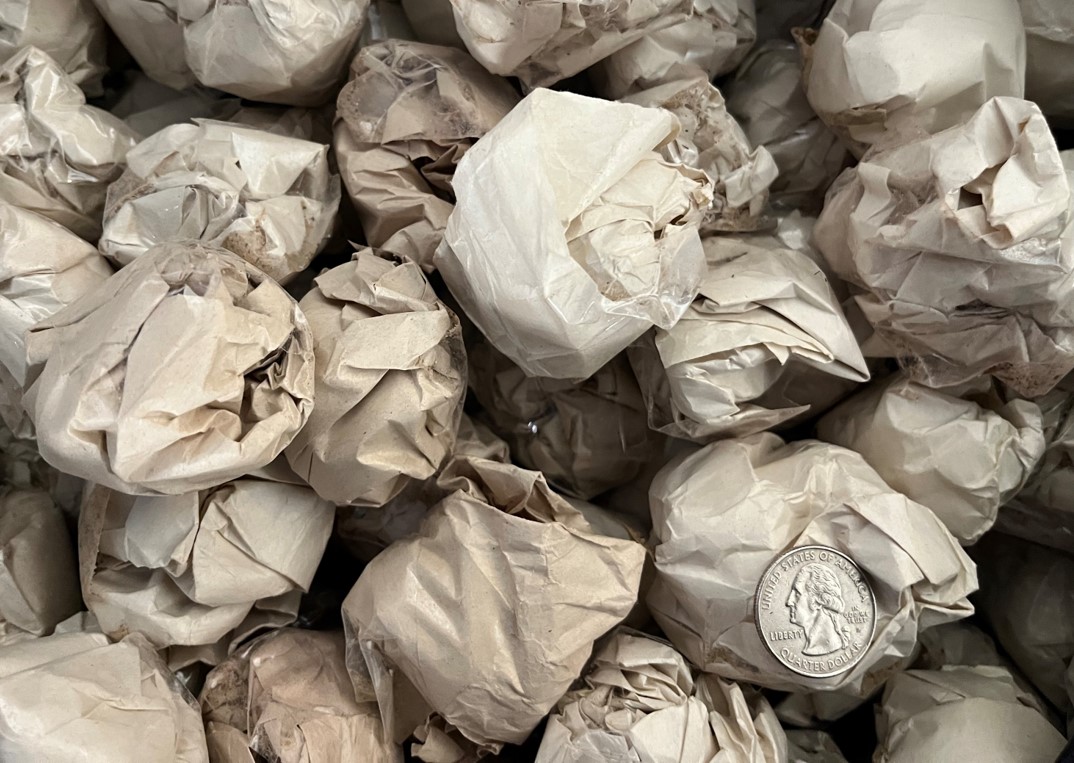} & 
     \includegraphics[width=0.146\textwidth]{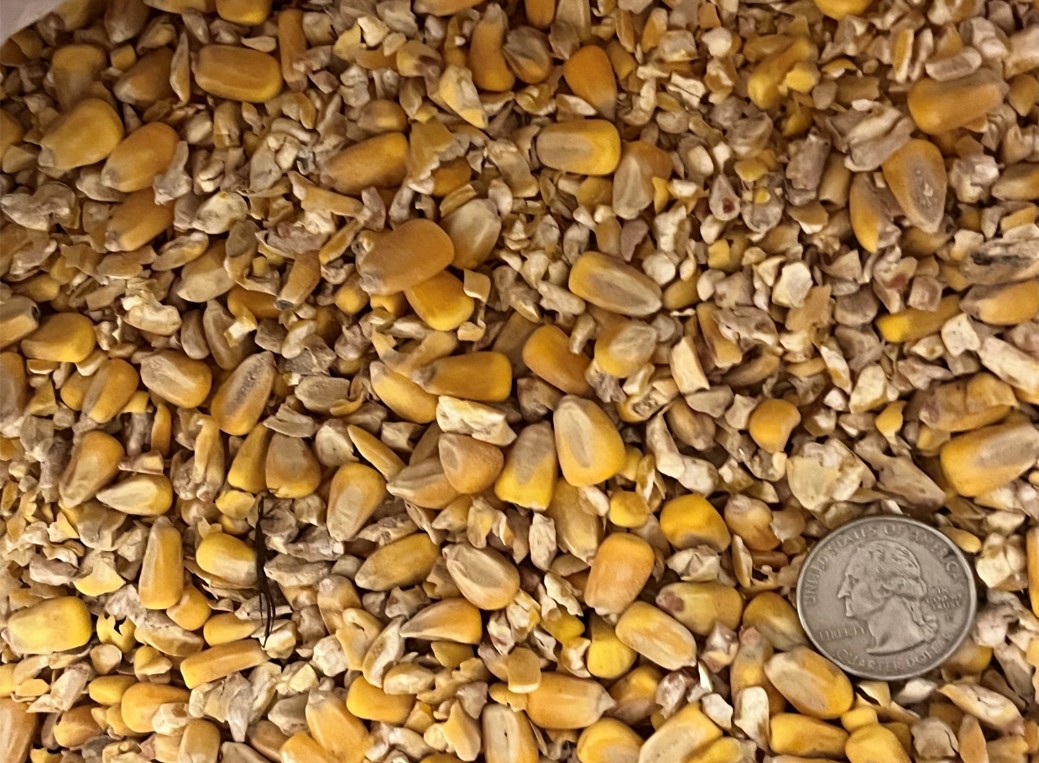} &
     \includegraphics[width=0.15\textwidth]{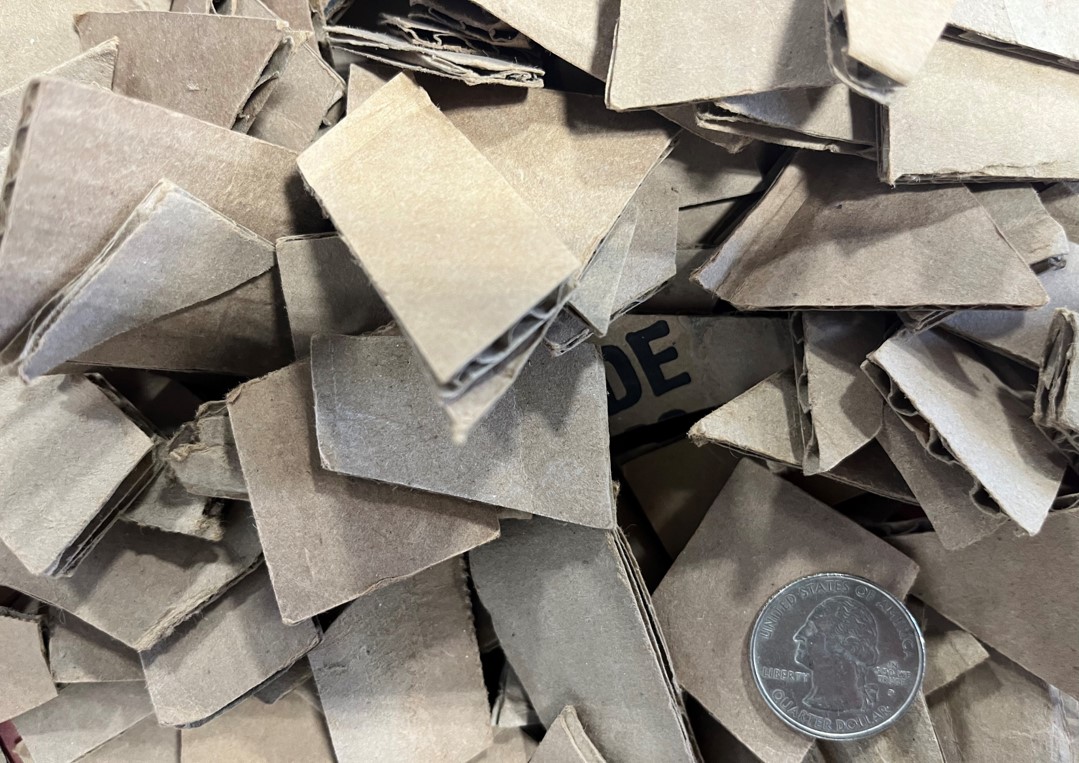} &
     \includegraphics[width=0.15\textwidth]{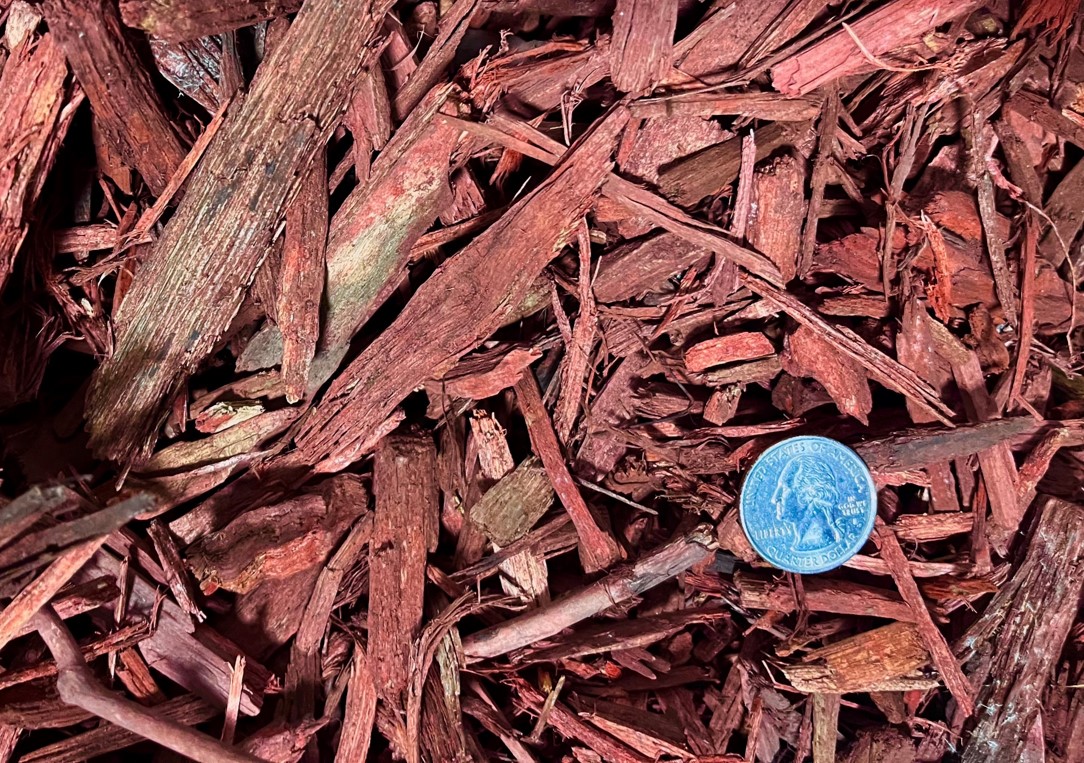} & 
     \includegraphics[width=0.15\textwidth]{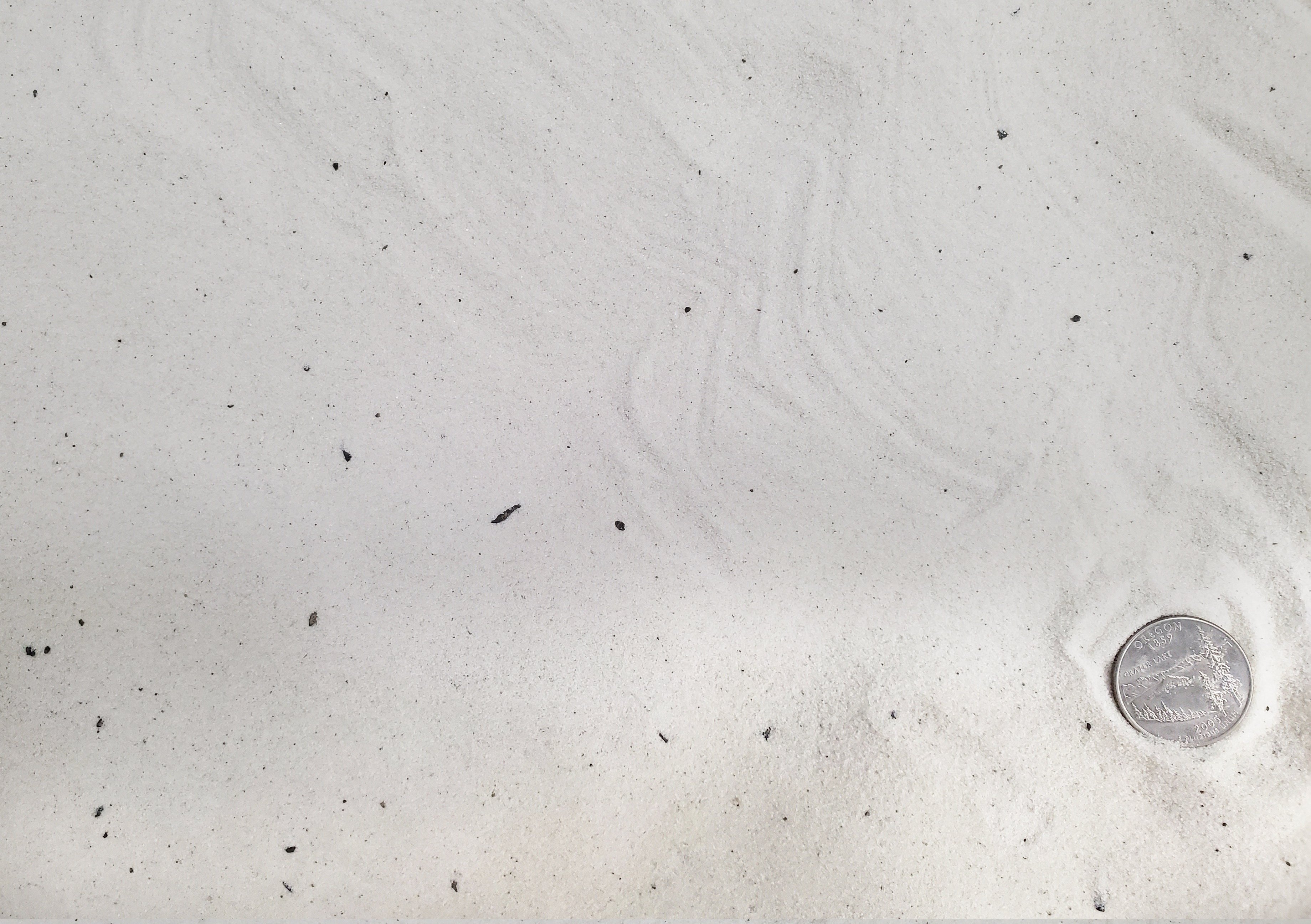} \\
     \bottomrule
     \end{tabular}
     \end{center}
 \end{table}

 Following the experimental procedure in \cite{zhu2023codega}, we compared the proposed method to the non-adaptive baseline (Non-Adaptive), i.e., only the deep mean component of the CoDeGa-trained model, \bt{which is a neural network that predicts the mean value of the estimated scoop volume,} and a volume maximizing (Vol-Max) policy, where the action is chosen to maximize the intersection between the scoop’s swept volume and the terrain following a strategy proposed in the excavation literature \cite{yang2021optimization}.

\begin{figure}[h!]
    \centering
    \begin{subfigure}[b]{0.324\textwidth}
        \centering
        \includegraphics[width=\textwidth]{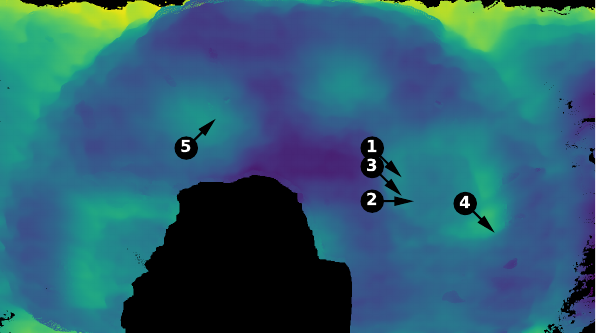}
        \caption{Vol-Max}
        \label{fig:volmax_results}
    \end{subfigure}
    \hfill
    \begin{subfigure}[b]{0.324\textwidth}
        \centering
        \includegraphics[width=\textwidth]{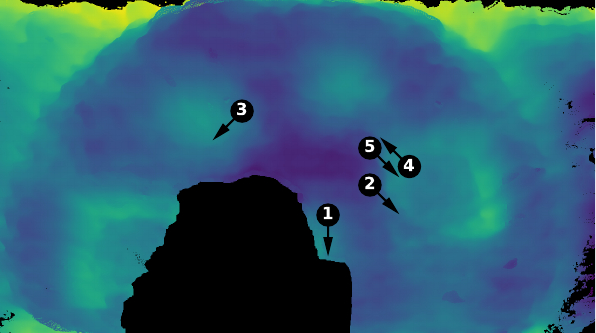}
        \caption{Non-Adaptive}
        \label{fig:nonadapt}
    \end{subfigure}
    \hfill
    \begin{subfigure}[b]{0.324\textwidth}
        \centering
        \includegraphics[width=\textwidth]{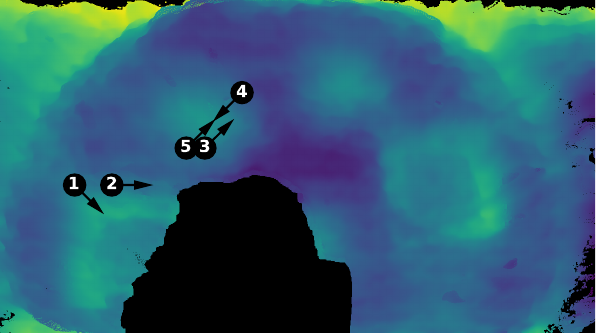}
        \caption{CoDeGa}
        \label{fig:codega_results}
    \end{subfigure}
    \caption{Example scooping attempts on the OWLAT testbed using three different methods. The cyan-to-yellow colors represent regions with higher terrain elevations.}
    \label{fig:three_methods}
    \end{figure}

 We conducted three iterations of the experiment using each method across three different scenarios, each featuring varying terrain topologies. Scenarios 1 and 2 consist of flat \texttt{Regolith} regions and \bt{higher} unscoopable \texttt{Comet} areas, with minor terrain features in the \texttt{Regolith} region that occur naturally during resets. In Scenario 3, the scoopable \texttt{Regolith} region is designed with terrain features that have heights comparable to those of the unscoopable \texttt{Comet} regions. The depth map in Fig. \ref{fig:three_methods} illustrates the terrain topology for Scenario 3, where the three circular green areas represent mounds created in the \texttt{Regolith} material.

 The objective is to maximize the total volume of material scooped in $k = 5$ attempts. Given that only one of the two materials in the testing terrain is scoopable, we measured mass in each attempt rather than volume to enhance measurement accuracy and ease. We used the average scooped mass over a run as the metric for comparing the efficacy of different methods. For all three scooping approaches, if the robot trajectory planning for a selected action during an attempt fails, the subsequent highest scoring action is selected until planning succeeds. We used the candidate action set outlined in Section \ref{sec:action_candidates}.

 \begin{table}
    \centering
    \caption{Results comparing CoDeGa with two baselines in OWLAT testbed. Higher scooped mass is better.}
    \begin{tabular}{lccc}
        \toprule
         & Vol-Max (gram) & Non-Adaptive (gram) & \textbf{CoDeGa (gram)} \\
        \midrule
        \textbf{Scenario 1} & 0.0 & 3.5 & \textbf{52.2} \\
        \textbf{Scenario 2} & 0.0 & 18.8 & \textbf{64.2} \\
        \textbf{Scenario 3} & 5.6 & 43.6 & \textbf{75.4} \\
        \midrule
        \textbf{Average} & 1.9 & 22.0 & \textbf{63.9} \\
        \bottomrule
    \end{tabular}
    \label{tbl:results}
\end{table}

The average scooped mass for all methods across the three different scenarios is reported in Table \ref{tbl:results}. \bt{The proposed approach that utilized CoDeGa-trained model} significantly outperformed the baseline methods in all scenarios. Fig. \ref{fig:three_methods} depicts the scooping actions executed during a representative trial for each method. The Vol-Max method consistently selected scoop locations near the \texttt{Comet} regions due to its inherent preference for areas with steep terrain gradients. While the Non-Adaptive baseline initially targeted the \texttt{Regolith} mounds, it failed to modify its policy in response to the data observed online, eventually resorting to ineffective scooping attempts in the \texttt{Comet} region, akin to Vol-Max. On the other hand, our approach initially engaged the \texttt{Comet} region but rapidly adapted its strategy in response to the low mass of scooped material, shifting focus to the \texttt{Regolith} mounds and thereby maximizing the total scooped mass.

\section{Conclusion}
This paper presents the deployment of a CoDeGa-trained adaptive scooping model on the high-fidelity OWLAT testbed. We evaluated the model, originally developed on a low-fidelity UIUC testbed, on novel terrains in OWLAT testbed with out-of-distribution materials. Experimental results demonstrate the model's capability to rapidly adapt to unfamiliar environments and make effective decisions despite significant domain shifts. Specifically, the model achieved several times higher average scooped mass compared to non-adaptive baselines across varying terrain topologies by adapting its strategy online based on limited experience.

The successful deployment substantiates the potential utility of learning-based autonomy for maximizing scientific return under uncertainty in ocean world missions. Moreover, it provides insights into real-world integration challenges, including data preprocessing and motion planning. Overall, this work endorses the feasibility of transferring learning-based systems from idealized training settings to realistic deployment environments. 

\section*{Acknowledgments}
This work was supported by NASA Grant 80NSSC21K1030. Part of this work was carried out at the Jet Propulsion Laboratory, California Institute of Technology, under a contract with the National Aeronautics and Space Administration (80NM0018D0004).

\bibliography{references}

\end{document}